\documentclass[11pt,a4paper]{article}
\usepackage[hyperref]{acl2021}
\usepackage{times}
\usepackage{latexsym}

\usepackage{microtype}
\usepackage{multirow}
\usepackage{graphicx}
\usepackage{booktabs}
\usepackage{amsmath}
\usepackage{amsfonts} 
\usepackage{xcolor}
\usepackage{float}
\usepackage{multirow}

\aclfinalcopy 

\title{Automatically Select Emotion for Response via \\Personality-affected Emotion Transition}

\author{Zhiyuan Wen, Jiannong Cao, Ruosong Yang, Shuaiqi Liu, Jiaxing Shen\\
  Department of Computing, \\The Hong Kong Polytechnic University\\ Kowloon, Hong Kong, China\\
  \texttt{\{cszwen, csjcao, csryang, cssliu, jiaxshen\}@comp.polyu.edu.hk} }

\date{}

\begin{document}
\maketitle
\begin{abstract}
To provide consistent emotional interaction with users, dialog systems should be capable to automatically select appropriate emotions for responses like humans. However, most existing works focus on rendering specified emotions in responses or empathetically respond to the emotion of users, yet the individual difference in emotion expression is overlooked. This may lead to inconsistent emotional expressions and disinterest users. To tackle this issue, we propose to equip the dialog system with personality and enable it to automatically select emotions in responses by simulating the emotion transition of humans in conversation. In detail, the emotion of the dialog system is transitioned from its preceding emotion in context. The transition is triggered by the preceding dialog context and affected by the specified personality trait. To achieve this, we first model the emotion transition in the dialog system as the variation between the preceding emotion and the response emotion in the \textbf{V}alence-\textbf{A}rousal-\textbf{D}ominance (VAD) emotion space. Then, we design neural networks to encode the preceding dialog context and the specified personality traits to compose the variation. Finally, the emotion for response is selected from the sum of the preceding emotion and the variation. We construct a dialog dataset with emotion and personality labels and conduct emotion prediction tasks for evaluation. Experimental results validate the effectiveness of the personality-affected emotion transition.\footnote{Our dataset is released at: github.com/preke/PELD}.
\end{abstract}

\section{Introduction}



Emotional intelligence can be considered a mental ability to reason validly with emotional information, and the action of emotions to enhance thought \cite{mayer20042004}. Hence, to create dialog systems with emotional intelligence during communication, it is necessary to enable the machine to understand the emotion of users, select appropriate response emotions and express in conversation.



Existing works either focus on rendering specified emotions in responses \cite{zhou2018emotional,colombo2019affect}, or understanding the emotion of users and respond empathetically \cite{zandie2020emptransfo,zhong2020endowing,lin-etal-2019-moel}; but how to automatically select the emotion for response is seldom discussed. \citet{wei2019emotion} proposes to learn appropriate emotional responses from massive anonymous online dialogues. However, trained on conversations from different speakers, the dialog system ignores the individual difference of expressing emotions. This may lead to inconsistent emotional interactions and disinterest users as they may feel they are still talking to rigid machines. 



In a dialog system, automatically selecting the emotion for response is to decide an emotion to be expressed facilitating the emotional response generation. Emotion selection can be modeled as the emotion transition \cite{thornton2017mental}, which refers to how the preceding emotion changes to the next, of the dialog system reacting to the dialog context. To achieve it like humans, it requires long-term patterns of thought, and behavior associated with an individual \cite{ball2000emotion}. \citet{mehrabian1996analysis} shows that the personality, \textit{e.g.}, the big-five personality model \cite{costa1992normal} also can be represented as temperament in the \textbf{V}alence-\textbf{A}rousal-\textbf{D}ominance (VAD) space for emotions \cite{mehrabian1996pleasure}.\footnote{It is Pleasure-Arousability-Dominance (PAD) in the original paper, PAD and VAD share the same meaning in the context of text understanding, we will use VAD for consistency henceforth.} The finding suggests that different personalities make different impacts on emotional expressions. Inspired by these works, we propose a personality-affected emotion transition model to endow personality to the dialog system, enabling it to select emotions that react to the dialog context affected by its given personality. 

In our method, we model the emotion transition of the dialog system as the variation in the VAD space from its preceding emotion to the next emotion in the response to users. We first obtain the preceding emotion of the dialog system from the dialog context and project it into the VAD space as an emotion vector. Simultaneously, we endow a personality trait, a 5-dimension vector representing the strength of each dimension in the big-five personality traits, to the dialog system. Then, we design neural networks to encode the dialog context and the personality traits into the VAD space to compose the variation of emotion. Finally, the emotion for response is selected based on the sum of the preceding emotion and the variation.

To facilitate related researches, we construct the \textbf{P}ersonality \textbf{E}motion\textbf{L}ines \textbf{D}ataset (PELD), which includes 6,510 dialogue triples of daily conversations with emotion labels and annotated personality traits. The emotion labels and personality annotations are adopted from other researches \cite{poria2018meld,zahiri2017emotion,jiang2019automatic} analyzing the script of a famous TV series \textit{Friends}\footnote{https://en.wikipedia.org/wiki/Friends}. We conduct emotion prediction tasks on the PELD dataset to evaluate the effectiveness of our method. The results suggest that the personality-affected emotion transition does contribute to better accuracy in emotion selection. To summary, our contributions are as follows:
\begin{itemize}
\item We raise the problem of automatically select the emotion for response in conversation and propose a new perspective to solve it through personality-affected emotion transition.
\item We construct a dialog script dataset with emotion and personality labels and analyze the patterns of emotion transitions in our dataset to facilitate related researches.
\item We evaluate the effectiveness of our proposed method on emotion prediction tasks and analyze the effects of personality and emotion transition  respectively.
\end{itemize}

\section{Related Works}
Our research is related to the emotional dialog systems, and the personality influence on emotion expression in psychology and Human-Computer Interaction (HCI). So, we review existing works in the two aspects as follows.
\subsection{Emotional Dialog Systems}
The concept of the emotional dialog system first occurred in \cite{colby1975artificial}, where a rule-based emotion simulation chatbot was proposed. Microsoft introduced the Xiaoice \cite{zhou2020design}, an empathetic social chatbot that is able to recognize users’ emotional needs, in 2014. Related researches become popular recently since \citet{zhou2018emotional} proposed the Emotional Chatting Machine to exploit the deep learning approach in building a large-scale emotionally aware conversational bot. Most existing works focus on incorporating specified emotion factors into neural response generation. \citet{shantala2018neural} trains emotional embeddings based on context and then integrated them into response generation. \citet{colombo2019affect} controls the emotional response generation with both categorical emotion representations and continuous word representations in VAD space \cite{mohammad2018obtaining}. Moreover, \citet{asghar2018affective} proposes an affectively diverse beam search for decoding. Besides, reinforcement learning is also adopted to encourage response generation models to render specified emotions. \citet{li2019reinforcement} combines reinforcement learning with emotional editing constraints to generate meaningful and customizable emotional replies. \cite{sun2018emotional} also uses an emotion tag to partially rewarding the model to express specified emotion.

However, it is impractical to always specify response emotions for dialog systems in real application scenarios. To simulate the emotional interaction among humans, \citet{wei2019emotion} designs an emotion selector to learn the proper emotion for responses from massive dialogue pairs. But the emotional expression is subjective, for the same post, different users may have different emotions in their responses. So, the pattern learned only from online dialogues ignores the user information and turns to be impractical. 
\subsection{Personality Effects on Emotions}


Emotion is a complex psychological experience of an individual’s state of mind as interacting with people or environmental influences \cite{han2012robotic}. The \textbf{P}leasure-\textbf{A}rousal-\textbf{D}ominance (PAD) \cite{mehrabian1996pleasure} or \textbf{V}alence-\textbf{A}rousal-\textbf{D}ominance (VAD) emotion temperament model shows three nearly orthogonal dimensions providing a comprehensive description of emotional states. Based on this, several psychologists studied the relationship between human emotional factors and personality factors. However, most of them are rule-based models \cite {johns2001emotions} and probabilistic models \cite{andre1999integrating}. \citet{mehrabian1996analysis} utilized the five factors of personality  \cite{costa1992normal} to represent the VAD temperament model through linear regression analysis. This finding is widely used to design robots having non-verbal emotional interaction with users \cite{han2012robotic,masuyama2018personality}, where the pre-defined personalities of robots affect their propensity of simulated emotion transitions. 

To integrate the analysis above into Artificial Intelligence, some researchers in HCI borrow the idea and design facial emotional expressions for humanoid robots. \citet{ball2000emotion} utilizes models of emotions and personality encoded as Bayesian networks to generate empathetic behaviors or speech responses to users in conversation. \citet{han2012robotic} employed five factors of personality to a 2D (pleasure-arousal) scaling model to represent a robotic emotional model. \citet{masuyama2018personality} introduces an emotion-affected associative memory model for robots expressing emotions. While in NLP, though the VAD space is adopted to model emotions in some researches \cite{mohammad2018obtaining,colombo2019affect, asghar2018affective}, the personality influence on emotion in dialogues is still an open problem.


\section{Methodology}

\subsection{Problem Definition}

We research on enabling the dialog system to automatically select emotions for response through the personality-affected emotion transition. 

Formally, a dyadic emotional conversation between the user and the dialog system contains the dialog context \textbf{$C=\{U_1, U_2, ..., U_{n-1}\}$} including all the preceding $n-1$ utterances from both the user and the dialog system, the preceding emotion $E_i$ expressed in $U_i \in C$ which is the last utterance from the dialog system, and the response emotion $E_r$ for the dialog system to facilitate generating the next emotional response $U_n$ to the user. We specify a personality trait $P_n$ to the dialog system and enable it to select response emotion $E_r$  through the personality-affected emotion transition model $F_{ET}$:

\begin{equation}
	\begin{aligned}
	E_r = F_{ET}(E_i|P_n, C)
	\end{aligned}
	\label{problem_equ}
\end{equation}

\noindent
where $E_r$ is transitioned from $E_i$. The transition is triggered by the preceding dialog context $C$ and affected by the specified personality trait  $P_n$. In the following content, we will introduce how we model this process in detail.

\begin{figure*}[t]
\centering
\includegraphics[scale=0.45]{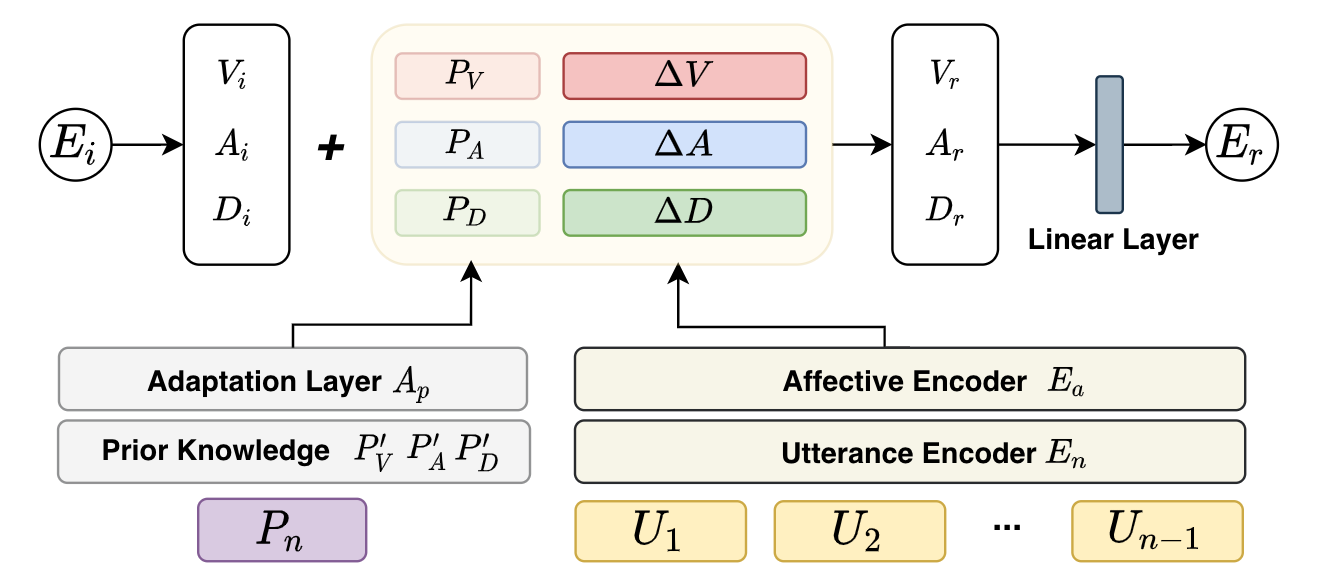} 
\caption{The Model Illustration}
\label{model}
\end{figure*}

\subsection{Preliminaries}


\subsubsection{Emotions in the VAD space}

Assuming in the problem above, emotions in all emotional utterances can be categorized into the six basic emotions: \textit{Anger}, \textit{Disgust}, \textit{Fear}, \textit{Joy}, \textit{Sadness}, and \textit{Surprise} \cite{ekman1994nature}. We project the basic emotions into the Valence-Arousal-Dominance (VAD) space as Table \ref{Emo_mapping} refer to the analysis result in \cite{russell1977evidence}\footnote{\textit{fear} and \textit{Joy} correspond to \textit{Terrified} and \textit{Happy} in the reference table.}. The VAD space indicates emotion intensity in three different dimensions, where the valence measures the positivity/negativity, arousal the excitement/calmness, and dominance the powerfulness/weakness. As for the utterances with no explicit emotion, we use the \textit{Neutral} with (0.00, 0.00, 0.00) as the VAD vector.

\begin{table}[h]
    \centering
    \linespread{1.4}
    \small
    \begin{tabular}{c|c}
        \toprule  
        \textbf{Basic Emotions} & \textbf{(Valence, Arousal, Dominance)}\\
        \hline  
		Anger & (-0.51, 0.59, 0.25) \\		
		Disgust & (-0.60, 0.35, 0.11) \\
		Fear & (-0.62, 0.82, -0.43) \\
		Joy & (0.81, 0.51, 0.46)\\
		Neutral & (0.00, 0.00, 0.00) \\
		Sadness & (-0.63, -0.27, -0.33) \\
		Surprise & (0.40, 0.67, -0.13) \\
		\bottomrule
    \end{tabular}
    \caption{Emotions in the VAD Space.}
    \label{Emo_mapping}
\end{table}

\subsubsection{Personalities in the VAD space}

Meanwhile, the big-five personality traits (OCEAN, shown in Table \ref{ocean traits}) are widely used for psychological analysis. \citet{mehrabian1996analysis} proposed a temperament model shown in Equation \ref{personality} derived through linear regression to show the VAD scales of personality traits, where $O,C,E,A,N$ are the strength of the big-five personality traits.

\begin{table}[b]
    \centering
	\scriptsize
    \begin{tabular}{ll}
        \toprule  
        \textbf{Factor} & \textbf{Description}\\
        \midrule  
		Openness & Openminded, imaginative, and sensitive. \\		
		Conscientiousness & Scrupulous, well-organized.\\
		Extraversion & The tendency to experience positive emotions. \\
		Agreeableness & Trusting, sympathetic, and cooperative. \\
		Neuroticism	&  The tendency to experience psychological distress. \\
		\bottomrule
    \end{tabular}
    \caption{The OCEAN personality traits and description \cite{costa1992normal}}
    \label{ocean traits}
\end{table}

\begin{equation}
	\begin{aligned}
	P_V &= 0.21E + 0.59A + 0.19N \\
	P_A &= 0.15O + 0.30A - 0.57N \\
	P_D &= 0.25O + 0.17C + 0.60E - 0.32A \\
	\end{aligned}
	\label{personality}
\end{equation}

\subsection{Personality-affected Emotion Transition}

Based on the problem definition and the preliminaries above, we design the Personality-affected Emotion Transition model as illustrated in Figure \ref{model}. Our model mainly include three modules: the personality effect on emotions in the left lower part, the context encoding in the right lower part, and the emotion transition in the top half in Figure \ref{model}. We will introduce these three modules in detail as follow.

\subsubsection{Personality Effect on Emotions} 
In our model, the personality of the dialog system is specified as a 5-dimensional vector $P_n = [O,C,E,A,N]$ representing the strength in Openness, Conscientiousness, Extraversion, Agreeableness, and Neuroticism, respectively.

The temperament of personality in the VAD space (shown in Equation \ref{personality}) is widely used as weighting parameters for emotion transition of robots in HCI works \cite{han2012robotic,masuyama2018personality}. However, the numeric coefficients in Equation \ref{personality} are summarized from analysis of questionnaire results from 72 participants \cite{mehrabian1996analysis}, which are not suitable to directly adopted as hyper-parameters in the model design. Hence, we choose to adopt the analysis results in Equation \ref{personality} as prior knowledge and learn suitable coefficients for personality by neural networks. First, we still calculate $P^\prime_V, P^\prime_A, P^\prime_D$ from the personality $P_n$ by Equation \ref{personality}; then we use $P^\prime_V, P^\prime_A, P^\prime_D$ as initialized input for an adaptation layer $A_p$ to learn the weighting parameters $P_V, P_A, P_D$ that suitable for the training data.

\subsubsection{Context Encoding} 

The dialog context acts as a set of parameters that may influence a person to speak an utterance while expressing a certain emotion \cite{poria2018meld}. In the VAD space, the emotion transition is regarded as the variation from one point (the preceding emotion) to another point (the next emotion). Thus, we generate the emotion transition variations $\Delta V, \Delta A, \Delta D$ from the semantic representations of the preceding dialog context $C$.

\begin{equation}
\begin{aligned}
\small
& R_{c} = E_n(U_1) \oplus E_n(U_2)  ... \oplus E_n(U_{n-1}) \\
& \Delta V, \Delta A, \Delta D = E_{a}(R_{c}) \\
\end{aligned}
\label{context}
\end{equation}

We fine-tune the pre-trained RoBERTa\footnote{Here we adopt the pre-trained RoBERTa-base model.} \cite{liu2019roberta} encoder, a famous pre-trained language model whose performance is widely validated in many natural language understanding tasks, to first extract the semantic representations $E_n(U_1), ..., E_n(U_{n-1})$ of all $n-1$ utterances in $C$. Then, we concatenate the semantic representations of utterances to obtain the overall context semantics $R_c$. Finally, $\Delta V, \Delta A, \Delta D$ are calculated by feeding $R_c$ into an affective encoder $E_{a}$, which extract the affective information from $R_c$ in the aspect of $V, A, D$, respectively.

\subsubsection{Emotion Transition} 

After we obtain the weighting parameters $P_V, P_A, P_D$ and the emotion transition variation $\Delta V, \Delta A, \Delta D$, the emotion for response is generated by the sum of the VAD vectors of the preceding emotion and the weighted variation, as shown in Equation \ref{emo_tran}.

\begin{equation}
\begin{aligned}
V_r &= V_i + P_V \cdot \Delta V\\
A_r &= A_i + P_A \cdot \Delta A\\
D_r &= D_i + P_D \cdot \Delta D\\
E_r &= F_c(V_r, A_r, D_r) \\
\end{aligned}
\label{emo_tran}
\end{equation}

\noindent
where the $V_i, A_i, D_i$ are the VAD vectors of $E_i$, and the $V_r, A_r, D_r$ are the emotion transition results in the VAD space. To alleviate the errors of using the numeric value in calculated VAD vectors, we add a linear layer $F_c$ to transform $V_r, A_r, D_r$ into a probability distribution on the discrete emotion categories. The output $E_r$ is the emotion with the largest probability.






\section{The PELD Dataset}

\subsection{Dataset Construction \& Statistics}

To facilitate related researches, we construct the \textbf{P}ersonality \textbf{E}motion\textbf{L}ines \textbf{D}ataset (\textbf{PELD}), an emotional dialog dataset with personality traits for speakers. As labeling online conversation on social media with speakers' personalities is time-consuming and may cause privacy issues, we turn to research on the dialogue script of a famous TV series \textit{Friends}. This classic script is widely analyzed in many dialog researches \cite{li2016persona,li2020transformers,jiang2019automatic}.

\begin{figure}[t]
\centering
\includegraphics[scale=0.24]{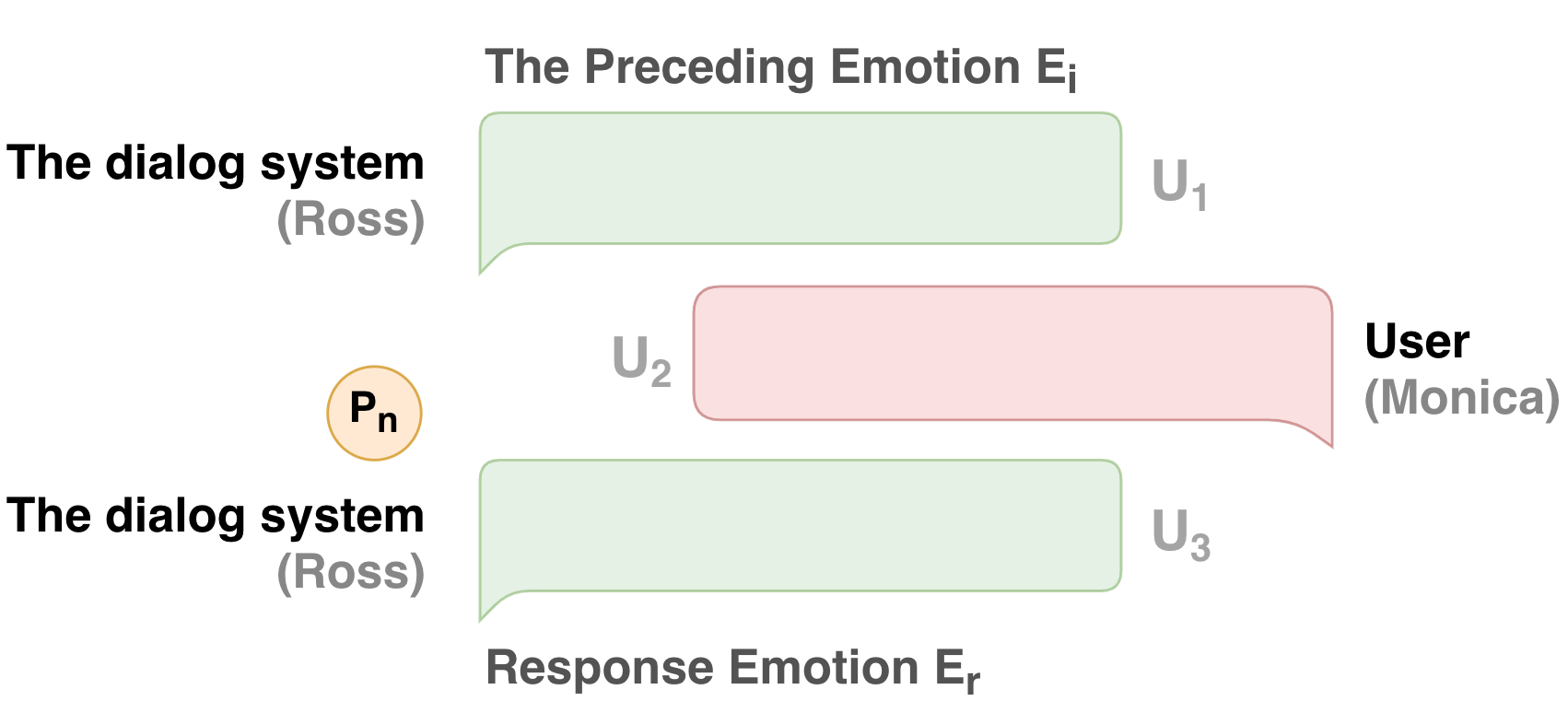} 
\caption{A triple example in PELD. The dyadic conversation between Ross and Monica (two main roles in \textit{Friends}, $P_n$ is the personality of Ross. The dialog system is set as Ross and talk with the user set as Monica in this example.}
\label{triple}
\end{figure}

In PELD, each sample is represented as a dialog triple ($C=\{U_1, U_2, U_3\}, \{E_i, E_r\}, P_n$), shown in Figure \ref{triple}) as a dyadic conversation. $E_i$ and $E_r$ are emotions expressed in $U_1$ and $U_3$, respectively. The utterances and their emotion labels are mainly adopted from the dialogues in the MELD \cite{poria2018meld} and the EmoryNLP dataset \cite{zahiri2017emotion}, two famous datasets analyzing emotional expressions in \textit{Friends}. To keep consistency, each dialog triple in PELD is constructed within the same dialogue in the original datasets. 

The personality traits in our dataset are adopted from the personality annotations in 711 different dialogues \cite{jiang2019automatic}. Refer to the annotations, a role may exhibit different aspects of its personality in different dialogues. We only keep the personality traits of the six main roles in \textit{Friends} for confidence as these annotations are most frequent. For each of the main roles, we average their annotated personality traits in all the dialogues by $P_n = \frac{1}{K}\sum_{i=1}^K{P_i}$ for simplification, where $K$ is the number of annotations. The averaged results are shown in Table \ref{personality_}.

\begin{table}[t]
    \centering
    \linespread{1.4}
    \small
    \begin{tabular}{l|l}
        \toprule  
       \textbf{Roles} & \textbf{Personality Traits (O,C,E,A,N)}\\
        \hline
		Chandler & [0.648, 0.375, 0.386, 0.58, 0.477]\\
		Joey & [0.574, 0.614, 0.297, 0.545, 0.455]\\
		Monica & [0.713, 0.457, 0.457, 0.66, 0.511]\\
		Phoebe & [0.6, 0.48, 0.31, 0.46, 0.56]\\
		Rachel & [0.635, 0.354, 0.521, 0.552, 0.469]\\
		Ross & [0.722, 0.489, 0.6, 0.533, 0.356]\\
		\bottomrule
    \end{tabular}
    \caption{Personalities of \textit{Friends} main roles in PELD.}
    \label{personality_}
\end{table}

 We split the PELD into \textbf{Train}, \textbf{Valid}, and \textbf{Test} set with portion around 8:1:1. The total number of utterances in PELD (10,648) is less than the sum of the original MELD (13,708) and the EmoryNLP (9,489), as not all dialogues are suitable to construct triples including main roles. The overall statistics of the dataset is shown in Table \ref{PELD_dataset}.

Similar to existing emotional conversation datasets \cite{li2017dailydialog, busso2008iemocap}, PELD also suffers the emotion imbalance issue. Utterances labeled as \textit{Neutral} are the majority, while \textit{Fear} and \textit{Disgust} only take a small portion. Though it reflects the real emotion distribution in daily conversation, it also brings challenges to machine learning models to identify and generate emotions. We tried several automatic methods for data augmentation like synonym substitution, back-translation, or the EDA proposed in \cite{wei2019eda}. But most of the synthetic samples are either odd or the same as the original samples. The reason might be there are limited options for short sentences as utterances in conversation to replace synonyms, add or delete words.

\begin{table}[t]
    \centering
    \linespread{1.4}
    \small
    \begin{tabular}{l|cccc} 
        \toprule
        \textbf{Basic Statistics} &\textbf{Train} & \textbf{Valid} & \textbf{Test} & \textbf{Total}\\		\hline
		 \#Triple & 5273 & 586 & 651 & 6510 \\
		 \#Unique Uttr. & 9306 & 1518 & 1675 & 10468\\
		 Avg. Uttr. Length & 9.26 & 9.33 & 8.95 & 9.32\\
		 \hline
		\textbf{\#Emotion} & \textbf{Train} & \textbf{Valid} & \textbf{Test} & \textbf{Total}\\
		\hline
		 Anger & 1863 & 236 & 241 & 2340\\
		 Disgust & 312 & 32 & 32 & 376\\
		 Fear & 1101 & 114 & 131 & 1346\\
		 Joy & 2863 & 326 & 344 & 3533 \\
		 Neutral & 7055 & 756 & 890 & 8701 \\
		 Sadness & 1088 & 121 & 136 & 1345\\
		 Surprise & 1537 & 173 & 179 & 1889\\
		\hline
		\textbf{\#Sentiment} & \textbf{Train} & \textbf{Valid} & \textbf{Test} & \textbf{Total}\\
		\hline
		 Positive & 4400 & 499 & 523 & 5422\\
		 Neutral & 7055 & 756 & 890 & 8701\\
		 Negative & 4364 & 503 & 540 & 5407\\
		\hline
		\textbf{\#Triple of Main Roles} & \textbf{Train} & \textbf{Valid} & \textbf{Test} & \textbf{Total}\\
		\hline
		 Chandler & 880 &  97 & 108 & 1085 \\
		 Joey & 912 & 109 & 102 & 1123 \\
		 Monica & 850 & 94 & 107 & 1051 \\
		 Phoebe & 782 & 87 & 103 & 972 \\
		 Rachel & 921 & 112 & 123 & 1156 \\
		 Ross & 928 &  87 & 108 & 1123 \\
		\bottomrule
    \end{tabular}
    \caption{Basic Statistics in PELD.}
    \label{PELD_dataset}
\end{table}

Another way to alleviate the imbalance issue is to expand the granularity of emotion to sentiment. As mentioned in \textbf{3.2}, in the VAD space, the Valence dimension of emotions measures the positivity and negativity, we can categorize the emotions into sentiments according to the values of Valance; i.e., positive emotions: \textit{Joy} and \textit{Surprise}; negative emotions: \textit{Anger}, \textit{Disgust}, \textit{Fear}, and \textit{Sadness}.  Thus, the distribution of sentiments in PELD is also shown in Table \ref{PELD_dataset}. Besides, dialog triples of six main roles (each triple corresponds to a main role with the personality trait) are averagely distributed in all train, valid, and test sets in PELD.




\subsection{Emotion Transitions in PELD}

After constructed PELD, we further explore the dataset in the aspect of emotion transitions. As the triples in PELD are constructed for analyzing the emotion transitions between $E_i$ in $U_1$ and the $E_r$ in $U_3$. Table \ref{Emo_Utter} shows the emotion and sentiment distributions in the $U_1$ and $U_3$, respectively. Besides, we also count the sentiments of emotions in $U_1$ and $U_3$ denoted as $S_1$ and $S_3$. We can see that for both emotion and sentiment, the distributions in $U_1$ and $U_3$ are similar, which means the transition of emotions and sentiments are equitable in PELD triples. Besides, the proportions of all emotions and sentiments are also similar to the overall statistics of PELD, which suggests that the emotions and sentiments in PELD are also average distributed in the triples.

\begin{figure*}[t]
\centering
\includegraphics[scale=0.41]{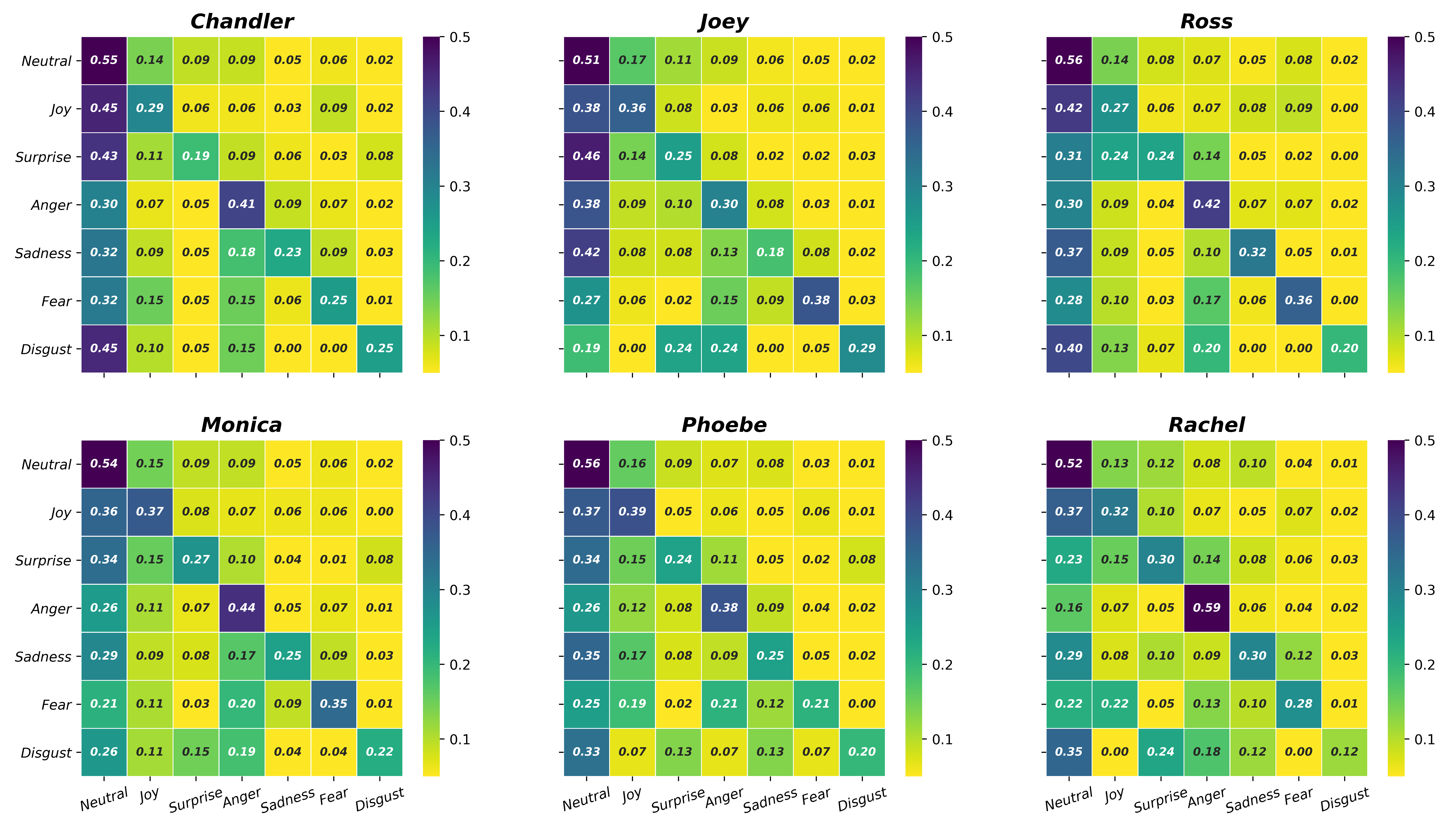}
\caption{Emotion transition matrixes of the six main roles in PELD. Each row in a matrix shows the ratios of the current emotion $E_i$ is transferred to the next emotion $E_r$.}
\label{Emotion_transition}
\end{figure*}

Since emotion transitions are affected by the personality traits as discussed above, we exhibit the emotion transition patterns for different roles with different personality traits in Figure \ref{Emotion_transition}. Although the emotion transitions are also correlated to the dialog context, we can still find patterns through these transition matrixes\footnote{Here, we analyze the personality-affected emotion transition based on roles rather than the numeric traits in Table \ref{personality} to avoid numeric observation errors.}.

In general, among the six transition matrixes, all the first columns are in deeper colors, which indicates most transitions occur from other emotions to \textit{Neutral} as it is the majority emotion in PELD. Besides, blocks with deeper color also more likely to occur around or in the diagonals of the transition matrixes; it suggests the preceding emotions tend to transition to the same or similar emotions. As for individual roles, 0.59 of the \textit{Anger} from Rachel remains the same in dialog triples, while for other roles, most \textit{Anger} emotions are transferred to \textit{Neutral} and \textit{Anger}. Besides, most \textit{Surprise} from Ross transfers to the \textit{Neutral}, \textit{Joy}, and \textit{Surprise}, but most \textit{Surprise}s of the other five roles tend to transfer to only \textit{Surprise} and \textit{Neutral}.

\begin{figure}[t]
\centering
\includegraphics[trim=0 50 0 0, scale=0.35]{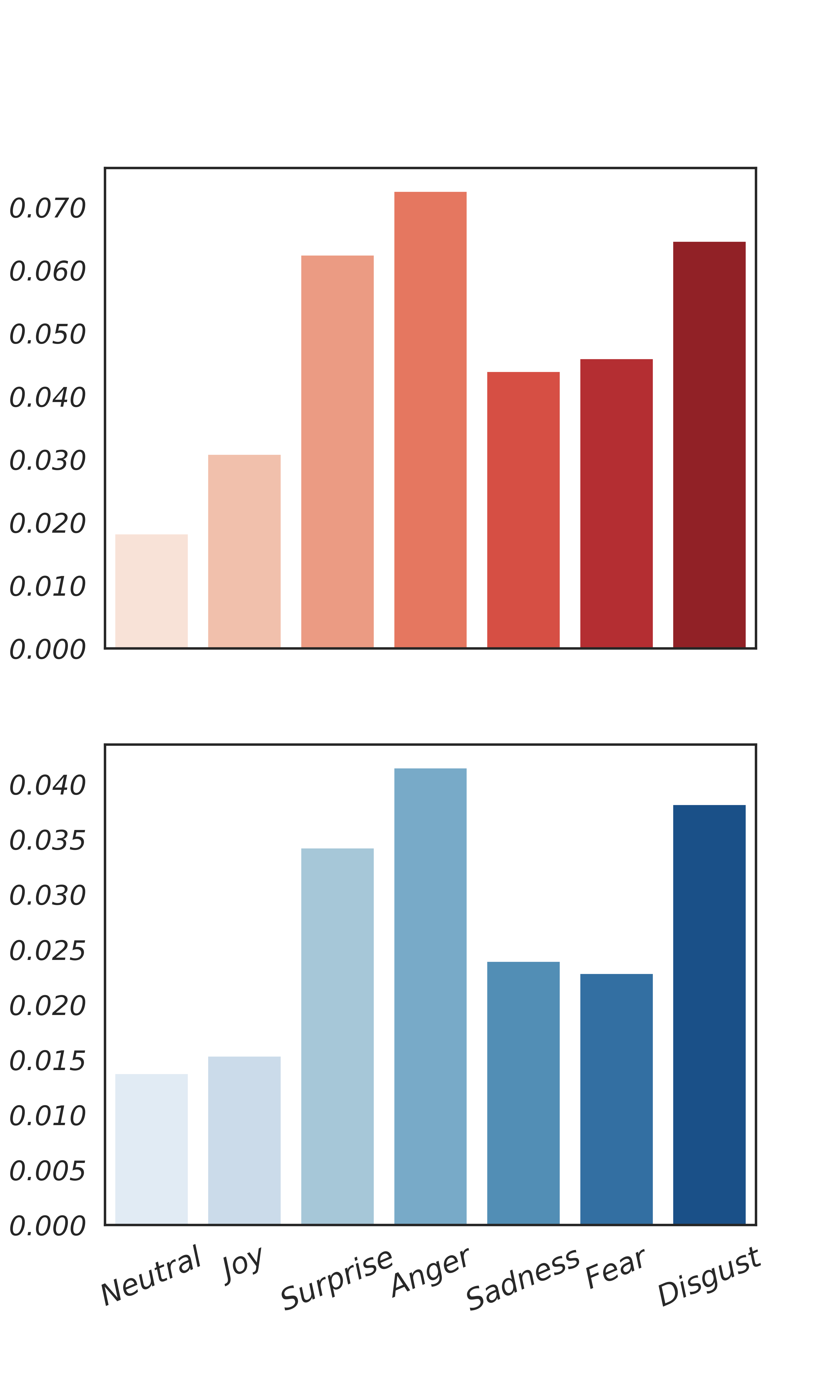}
\caption{The standard deviations of the infinity norm (red) and the L2-norm (blue) of each row in emotion transition matrixes of the six main roles in PELD.}
\label{Emotion_transition_all}
\end{figure}

Moreover, to highlight the individual differences of emotion transitions among the six main roles in detail, we also show the standard deviations (Std) of each row in the emotion transition matrixes of the six main roles, as shown in Figure \ref{Emotion_transition_all}. The red bar chart shows the Std of the infinite norms of rows in the emotion transition matrix, which indicates the diversity of the most probable emotions from the same emotion in emotion transfers of different roles. While the blue bar chart shows the Std of the L2-norms, which generally describes the difference in how different roles transfer from one emotion to other emotions.

Both charts show similar patterns of emotion transitions. \textit{Anger}, \textit{Surprise}, and \textit{Disgust} vary the most in different roles, while people are more common when process \textit{Neutral} and \textit{Joy} emotions in conversation. Besides, negative emotions (\textit{Anger}, \textit{Sadness}, \textit{Fear}, and \textit{Disgust}) are relatively higher than positive emotions and \textit{Neutral} on average. So, we can infer that the personality traits influence more in the emotion transfers from negative emotions.

%

\begin{table}[b]
    \centering
    \linespread{1.5}
    \setlength\tabcolsep{2pt}
    \scriptsize
    \begin{tabular}{c|c|c|c|c|c|c|c} 
        \toprule
		\textbf{Tri.Emos} & \textbf{Neutral} & \textbf{Joy} & \textbf{Surprise} & \textbf{Anger} & \textbf{Sadness} & \textbf{Fear} & \textbf{Disgust} \\
		\hline
		 $E_i$ & 2910 & 1242 & 597 & 751 & 438 & 457 & 115 \\
		 $E_r$ & 2771 & 1123 & 634 & 858 & 493 & 487 & 144 \\
		 \hline
		 \textbf{Tri.Sentis} & \textbf{Neutral} & \multicolumn{2}{c|}{\textbf{Positive}}&\multicolumn{4}{c}{\textbf{Negative}} \\
		 \hline
		 \textbf{ $S_i$ } & 2910 & \multicolumn{2}{c|}{1839}&\multicolumn{4}{c}{1761} \\
		 \textbf{ $S_r$ } & 2771 & \multicolumn{2}{c|}{1757}&\multicolumn{4}{c}{1982} \\
		\bottomrule
    \end{tabular}
    \caption{Emotions in PELD Triples}
    \label{Emo_Utter}
\end{table}

%



%

%

\section{Experiment}

\subsection{Evaluation Tasks}

To validate the effectiveness of our proposed emotion generation model, we set two evaluation tasks: Emotion Prediction and Sentiment Prediction on PELD. Emotion Prediction requires the model to predict the emotion in the upcoming utterance based on the preceding dialog context in a dyadic conversation scenario; while Sentiment Prediction has the same setting except to predict the sentiment in the upcoming utterance. 

For both tasks, we evaluate the prediction performance by F-scores of single emotion or sentiment. Besides, the overall performance is also measured from two aspects with the macro averaged (\textbf{m-avg}) and the weighted averaged (\textbf{w-avg}) F-scores. A higher m-avg indicates the model performs relatively better predicting all categories, while a higher w-avg indicates the model predicts emotions or sentiments with larger proportions in the dataset better.


\begin{table*}[t]
    \centering
    \linespread{1.4}
    \small
    \begin{tabular}{l|ccccccccc} 
        \toprule
        \textbf{Methods} & \textbf{Anger} & \textbf{Disgust} & \textbf{Fear} & \textbf{Joy} & \textbf{Neutral} & \textbf{Sadness} & \textbf{Surprise} & \textbf{m-avg} & \textbf{w-avg} \\
        		\hline
		 RoBERTa &0.218 & 0.000 & 0.107 & 0.214 & 0.453 & 0.122 & 0.126 & 0.177 & 0.287 \\
		 RoBERTa-P &0.178 & 0.000 & 0.047 & \textbf{0.265} & 0.517 & 0.110 & 0.053 & 0.167 & 0.352 \\
		 PET-VAD &0.190 & \textbf{0.081} & 0.115 & 0.188 & 0.474 & 0.000 & \textbf{0.179} & 0.175 & 0.309 \\
		 PET-CLS & \textbf{0.320} & 0.070 & \textbf{0.140} & 0.198 & \textbf{0.528} & \textbf{0.155} & 0.098 & \textbf{0.203} & \textbf{0.424} \\
		\bottomrule
    \end{tabular}
    \caption{Results for Emotion Prediction.}
    \label{Emotion_result}
\end{table*}

\subsection{Ablation Study Setting}


Although plenty methods  \cite{majumder2019dialoguernn,ghosal2020cosmic,ghosal2019dialoguegcn} has been proposed to analyze emotions in dialogues of \textit{Friends}, most of their targets are to recognize the emotions of utterances in conversation. Compared with emotion recognition, the problem setting of selecting emotion is different and it is more difficult to select the appropriate emotion in response without knowing the response content. So, instead of comparing with other emotion recognition models, we turn to conduct ablation studies to evaluate the effectiveness of different parts of our model design. The ablation study compares the performances of the following models:\\

\noindent
\textbf{RoBERTa:} RoBERTa \cite{liu2019roberta} is a famous pre-trained language model designed for natural language understanding. Its performance is widely validated in many downstream tasks. We here use pre-trained RoBERTa, corresponding to the $E_n$ in our model, to encode the preceding dialog context to obtain the semantic representation as input, then directly predict the emotion for response through a classification head. \\

\noindent
\textbf{RoBERTa-P:} We concatenate the personality vector of the speakers with the dialog context representation by RoBERTa as the feature, then predict the response emotion. This method is to evaluate whether personality influences the expression of emotions. \\

\noindent
\textbf{PET-VAD:} As emotions can be represented by both discrete category labels or vectors in the VAD spaces. PET-VAD is set to compare the different usages of emotion VAD vectors in our model. During training, PET-VAD regressions the VAD vectors of target emotions by minimizing the Mean Squared Error (MSE) between generated vectors and the VAD vectors of ground truth emotions. The prediction output of PET-VAD is the closest neighbor emotions of generated VAD vectors measured by MSE.\\


\noindent
\textbf{PET-CLS:} This is our method Personality-affected Emotion Transition with a classifier after obtaining the VAD vector of generated emotion. PET-CLS predicts emotions in the upcoming utterances as described in Section 3.\\

For RoBERTa, RoBERTa-P, and PET-CLS directly outputting discrete emotions, we adopt the Focal loss \cite{lin2017focal} to relieve the imbalanced emotion prediction.

\section{Results and Analysis}

In this section, we report and analyze the experimental results on the Test set of PELD in our ablation study. All results are chosen by the best performance on the Valid set within 50 epochs training. 

\subsection{Results for Emotion Prediction}

The results on the Emotion Prediction task are reported in Table \ref{Emotion_result}. First of all, as a seven-classes prediction task also suffered from the imbalance issue, the overall performance is moderately low, which also indicates the difficulty of the task. As for the averaged F-scores, PET-CLS improves both the w-avg and m-avg by a large margin from all other methods, which verifies our personality-affected emotion transition method.

In detail, all models perform better on emotions with larger portions (\textit{Neutral} and \textit{Joy}), as they are more probable to occur in the response emotion. Moreover, PET-VAD and PET-CLS achieve moderately higher F-score on the minority emotions (\textit{Anger}, \textit{Sadness}, \textit{Disgust}, \textit{Fear}, and \textit{Surprise}), which shows that the emotion transition process is more important generating these minority emotions. It also verifies the finding in Section 4.2.
 
On the other hand, although PET-VAD is based on the designed personality-affected emotion transition, most single emotion F-scores of PET-VAD are lower than RoBERTa or RoBERTa-P. We discuss the possible reasons as follows. One reason might be that the imbalance emotion issue cannot be alleviated in directly regression the emotion VAD vectors. Another reason might be that the value of emotion VAD vectors in Table \ref{Emo_mapping} are estimated rather than precisely calculated, and the distance among different emotions in the theoretical VAD space is not similar to those in the emotion distribution in daily conversation. 

\begin{table}[t] 
    \centering
    \linespread{1.5}
    \setlength\tabcolsep{2pt}
	\small
    \begin{tabular}{l|ccccc} 
        \toprule
        \textbf{Methods} &\textbf{Negative} & \textbf{Neutral} & \textbf{Positive} & \textbf{m-avg} & \textbf{w-avg}\\		\hline
		 RoBERTa & 0.415 & 0.430 & 0.323 & 0.389 & 0.390  \\
		 RoBERTa-P & 0.401 & \textbf{0.505} & 0.176 & 0.361 & 0.430 \\
		 PET-CLS & \textbf{0.492} & 0.474 & \textbf{0.327} & \textbf{0.431} & \textbf{0.445} \\
		\bottomrule
    \end{tabular}
    \caption{Results for Sentiment Prediction.}
    \label{Sentiment_result}
\end{table}
\subsection{Results for Sentiment Prediction}

As predicting the emotions for the upcoming responses is difficult due to the multiple imbalanced categories, we also report the results on the Sentiment Prediction task in Table \ref{Sentiment_result}. Besides, different from the analysis above, which categorizes emotions by their portions in PELD, sentiment is another aspect of emotion analysis. As the sentiments are not directly described in the VAD spaces, we only report the results for RoBERTa, RoBERTa-P, and the PET-CLS. Besides, we only change the output size of PET-CLS from 7 (for emotions) to (3 for sentiments) and preserve the emotion transition process in this task.

In general, we can see that the prediction F-scores of sentiments are higher than emotion predictions. Besides, the prediction of negative emotions is much easier than predicting positive emotions in all three methods. It may because although the numbers of sentiments are similar, the categories of negative emotions (\textit{Anger}, \textit{Sadness}, \textit{Fear}, and \textit{Disgust}) are more than positive emotions (\textit{Joy} and \textit{Surprise}). Equipped with our model design, PET-CLS outperforms both RoBERTa and RoBERTa-P excepted for the neutral sentiment. It suggests that the personality-affected emotion transitions also facilitate sentiment prediction. However, only concatenating the personality vectors with context representation, RoBERTa-P improves the F-scores of Neutral but decreases the Positive and Negative. Hence, direct concatenation limits the effect of personality information in sentiment prediction.

%

%
%

%
%
%
%



\section{Conclusion and Future Work}
In this work, we raise the problem of automatically selecting the emotion for response considering the individual differences in conversation and propose a new perspective to solve it through personality-affected emotion transition. Besides, we construct a dialog script dataset PELD with emotion and personality labels to facilitate related researches. We also validate our personality-affected emotion transition model in emotion prediction experiments.
 
Facial expressions, voices, gestures, and environment information are also vital in emotional interaction, but they are not captured in the purely text-based dialog systems. Besides, as seen from statistics in PELD, the most common emotion in the dialog scripts is still Neutral. One possible reason is that other subtle affective information is not captured in the text. Therefore, our future works will continue to investigate the personality effects on emotions in the multi-modality scenario. 



%
%
%
%
%
%
%
%
%
%
%
%
%
%
%
%
%
%
%
%
%
%
%

\subsection{Acknowledgement}
This work is supported by the Hong Kong RGC Collaborative Research Fund with project code C6030-18G and Hong Kong Red Swastika Society Tai Po Secondary School with project code P20-0021.
\bibliographystyle{acl_natbib}
\bibliography{anthology,acl2021}


\end{document}